\pdfoutput=1 %
\PassOptionsToPackage{table}{xcolor}
\PassOptionsToPackage{pagebackref,breaklinks,bookmarks=true}{hyperref}

\documentclass[sigconf, nonacm]{acmart}

\makeatletter
\def\@secfont{\sffamily\bfseries\Large\section@raggedright}
\def\@subsecfont{\sffamily\bfseries\large\section@raggedright}
\def\@subsubsecfont{\sffamily\itshape}
\def\@parfont{\sffamily\itshape}
\makeatother

\usepackage[utf8]{inputenc}   %
\usepackage[T1]{fontenc}      %
\usepackage[scaled]{helvet}   %

\usepackage{amsmath}
\usepackage{amssymb}
\usepackage{amsfonts}         %
\usepackage{bm}               %
\usepackage{nicefrac}         %
\usepackage[makeroom]{cancel} %

\usepackage{graphicx}
\usepackage{epsfig}           %
\usepackage[table]{xcolor}
\usepackage{tikz}
\usetikzlibrary{arrows.meta, backgrounds, calc, decorations.pathreplacing, fit, patterns}
\definecolor{ourportwine}{RGB}{128,25,55}
\usepackage{adjustbox}        %
\usepackage[most]{tcolorbox}  %

\usepackage{booktabs}         %
\usepackage{tabularx}

\usepackage{microtype}        %
\usepackage{url}              %
\usepackage{multicol}
\usepackage{dirtytalk}        %
\usepackage{soul}             %
\usepackage{epigraph}
\usepackage{comment}
\usepackage{xspace}
\usepackage{algorithm}
\usepackage{lipsum}

\usepackage[shortlabels]{enumitem}

\usepackage[font=small,labelfont=bf,textfont=normalfont]{caption}

\definecolor{ourblue}{rgb}{0.368,0.507,0.71}    %
\definecolor{ourdarkblue}{HTML}{354d72}         %
\definecolor{ourorange}{rgb}{0.881,0.611,0.142} %
\definecolor{ourgreen}{rgb}{0.56,0.692,0.195}   %
\definecolor{ourred}{rgb}{0.923,0.386,0.209}    %
\definecolor{ourviolet}{rgb}{0.528,0.471,0.701} %
\definecolor{ourbrown}{rgb}{0.772,0.432,0.102}  %
\definecolor{ourazure}{rgb}{0.364,0.619,0.782}  %
\definecolor{ourolive}{rgb}{0.572,0.586,0.}     %
\definecolor{ourgray}{RGB}{102,88,84}           %

\definecolor{ourblue2}{RGB}{9,134,223} %
\definecolor{ourdarkblue2}{RGB}{5,97,164} %
\definecolor{ourlightblue2}{RGB}{132,201,250} %

\definecolor{ourorange2}{RGB}{224,90,18} %
\definecolor{ourdarkorange2}{RGB}{160,63,9} %
\definecolor{ourlightorange2}{RGB}{246,175,137} %

\definecolor{ouryellow2}{RGB}{227,213,25} %
\definecolor{ourdarkyellow2}{RGB}{177,166,17} %
\definecolor{ourlightyellow2}{RGB}{242,235,140} %

\definecolor{ourpink2}{RGB}{247,24,139} %
\definecolor{ourdarkpink2}{RGB}{164,4,86} %
\definecolor{ourlightpink2}{RGB}{250,163,207} %

\definecolor{ourgreen2}{RGB}{159,198,52} %
\definecolor{ourdarkgreen2}{RGB}{109,138,30} %
\definecolor{ourlightgreen2}{RGB}{209,228,154} %

\definecolor{ourgray2}{RGB}{124,124,115} %
\definecolor{ourdarkgray2}{RGB}{87,87,81} %
\definecolor{ourlightgray2}{RGB}{194,194,189} %

\colorlet{primalconecolor}{ourblue2}        %
\colorlet{dualconecolor}{ourpink2}          %
\colorlet{polarconecolor}{ourgray2}         %
\colorlet{forcecolor}{ourportwine}          %
\colorlet{velocitycolor}{ourdarkgreen2}     %
\colorlet{correctioncolor}{ourorange}       %
\colorlet{stabilizationcolor}{ourviolet}    %
\colorlet{coordinatecolor}{ourdarkblue2}    %
\colorlet{torquecolor}{ourdarkorange2}      %
\colorlet{geometrycolor}{black}             %
\colorlet{bodycolor}{ourlightblue2}         %
\colorlet{fixedbodycolor}{ourlightgray2}    %
\colorlet{edgecolor}{ourdarkgray2}          %
\colorlet{annotationcolor}{ourdarkgray2}    %
\colorlet{procolor}{ourdarkgreen2}          %
\colorlet{concolor}{ourportwine}            %
\colorlet{relaxationcolor}{ourportwine}     %
\colorlet{highlightcolor}{ourblue2}         %
\colorlet{panelcolor}{ourblue!7}            %
\colorlet{mutedtextcolor}{ourgray2}         %
\colorlet{hyperlinkcolor}{ourblue}          %
\colorlet{biblinkcolor}{ourdarkblue}        %
\colorlet{bulletcolor}{ourblue}             %
\colorlet{boxcolor}{black!60}               %

\newtcolorbox{equationbox}[1]{
    enhanced,
    sharp corners,
    colback=white,
    colframe=boxcolor,
    boxrule=0.8pt,
    top=0.6em, %
    attach boxed title to top left={
        xshift=-1mm,
        yshift=-5mm,
        yshifttext=-1mm,
    },
    boxed title style={
        sharp corners,
        size=small,
        colback=boxcolor,
        colframe=boxcolor,
    },
    fonttitle=\bfseries,
    title=#1
}

\makeatletter
\newcommand*{\addFileDependency}[1]{%
  \typeout{(#1)}
  \@addtofilelist{#1}
  \IfFileExists{#1}{}{\typeout{No file #1.}}
}
\makeatother

\newcommand{\captitle}[1]{\textbf{#1}}

\makeatletter
\DeclareRobustCommand\onedot{\futurelet\@let@token\@onedot}
\def\@onedot{\ifx\@let@token.\else.\null\fi\xspace}
\makeatother

\newcommand{\trsp}{{\scriptscriptstyle\top}}
\newcommand{\Id}{\mathbf{I}}  %

\DeclareMathOperator*{\argmin}{arg\,min}

\input{contact_notation.tex}

\usepackage{algorithm}
\usepackage{algpseudocode}
\makeatletter
\renewcommand{\theHALG@line}{\thealgorithm.\arabic{ALG@line}}
\makeatother
\algrenewcommand\algorithmicrequire{\textbf{Input:}}
\newcolumntype{P}[1]{>{\raggedright\arraybackslash}p{#1}} %

\usepackage{etoolbox}
\newenvironment{mysubequations}
  {\begin{subequations}%
   }
  {\end{subequations}}

\newcommand{\code}[1]{\texttt{\detokenize{#1}}}
\newcommand{\aref}{a_{\text{ref}}}
\newcommand{\arefi}[1]{a_{\text{ref},#1}}
\newcommand{\arefp}[1]{a'_{\text{ref},#1}}

\AtEndPreamble{\hypersetup{colorlinks=true,allcolors=hyperlinkcolor}}
\usepackage[capitalise]{cleveref}
\usepackage{cinzel}
\DeclareRobustCommand{\Lodos}{%
  {\color{ourdarkblue2} \cinzel \selectfont LODOS-v2}%
}
\begin{document}

\title{\Lodos: An Efficient Primal Formulation of the Exact Frictional Contact Problem}

\author{Onur Beker}
\affiliation{\institution{University of T\"ubingen}\country{}}
\author{A.\ Ren\'e Geist}
\affiliation{\institution{University of T\"ubingen}\country{}}
\author{Georg Martius}
\affiliation{\institution{University of T\"ubingen}\country{}}

\begin{abstract}
Lodos-v2 is a GPU-based simulation framework for massively parallel rigid-body dynamics with contact. It is specifically designed to achieve high throughput in scenes involving collisions among large numbers of detailed meshes.
Its predecessor, Lodos-v1, had two main limitations. First, it solved the standard convex relaxation of cone-complementarity-based contact dynamics, thereby inheriting well-known physical artifacts, including lift-off and artificial compliance in the normal direction, as well as creep in the tangential direction. Second, its contact-group reparameterization scheme, which rescales spring-damper forces using a single factor per collision pair, could not be integrated with other constraints required by a full-fledged simulator, such as joint limits and welds.
Lodos-v2 addresses both limitations through two main contributions. First, it introduces an efficient algorithm for solving the exact frictional contact problem by sequentially solving a cascade of convex relaxations in primal form within an outer proximal minimization loop. Second, it extends the contact-group reparameterization scheme of Lodos-v1 to the primal formulation of cone-complementarity-based contact dynamics, enabling its integration with the other constraints of a full-fledged simulator.
Experiments demonstrate that these contributions result in throughput improvements over established rigid-body simulation frameworks. The source code is available at \href{https://github.com/martius-lab/lodos}{https://github.com/martius-lab/lodos}.
\end{abstract}

\begin{teaserfigure}
\includegraphics[width=\textwidth]{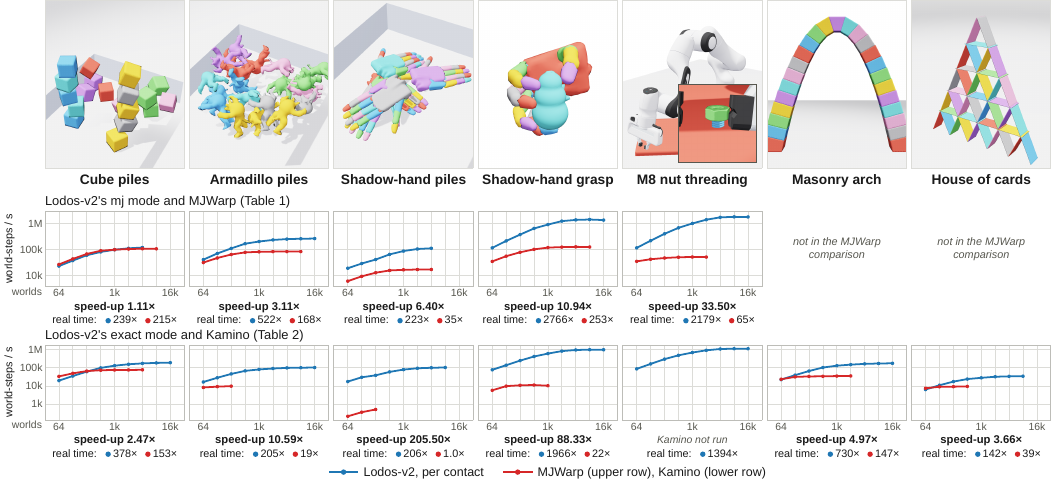}
\caption{\captitle{Benchmark scenes used to compare the throughput of Lodos-v2 against established simulation frameworks.} Top: A representative frame from each scene, simulated using Lodos-v2's exact mode. Bottom: The speed-up achieved by Lodos-v2 (using the per contact parameterization) over each baseline in terms of maximum achievable throughput, computed from \cref{tab:mjwarp,tab:kamino}.}
\label{fig:teaser}
\end{teaserfigure}

\maketitle

\section{Introduction}
Massively vectorized simulation of rigid-body dynamics with contact \cite{makoviychuk2021isaac, mujoco_warp, The_Newton_Contributors_Newton_GPU-accelerated_physics_2025} is essential for a range of robotics applications, including reinforcement learning (RL) \cite{hwangbo2019learning, allshire2022transferring}, sampling-based model-predictive control (MPC) \cite{kurtz2024hydrax, li2024_drop}, and randomized smoothing \cite{suh2022bundled, pang2023global} for gradient-based optimization. Lodos-v1~\cite{beker2026lodos} was developed to support such applications by enabling high-throughput simulation of scenes that contain large numbers of collision pairs between geometrically detailed meshes. Its two main contributions were (i) efficient broadphase and narrowphase collision routines for vertex-SDF contacts and (ii) a \say{contact-group reparameterization} scheme that scales spring-damper contact forces using a single factor per collision pair, thereby making the number of solver decision variables and constraints independent of the number of contacts. Lodos-v2 preserves the collision detection framework of Lodos-v1 while substantially extending its treatment of contact dynamics. In particular:
\begin{itemize}[leftmargin=*, label={\color{bulletcolor}\textbf{\textbullet}}]
    \item Lodos-v1 lacked a mechanism for integrating contact constraints with other constraint types commonly supported by general-purpose physics simulators, such as MuJoCo~\cite{todorov2012mujoco}, including joint limits, weld constraints, and friction-loss constraints. Lodos-v2 overcomes this limitation, enabling it to reproduce MuJoCo's contact dynamics at higher simulation throughput.
    \item Lodos-v1 solved the convex relaxation of cone-complementarity-based contact dynamics introduced by Anitescu et al.~\cite{anitescu2006optimization}, which is also employed by MuJoCo. As a result, it inherits the known physical artifacts of this relaxation, including sliding contacts lifting-off, normal-direction compliance, and tangential creep that prevents contacts from remaining stationary under zero velocity and acceleration. Lodos-v2 addresses these limitations by introducing a novel primal-form solver for the exact cone complementarity problem, building on the work of Carpentier et al.~\cite{carpentier2024compliant}. This enables Lodos-v2 to reproduce the contact dynamics of Kamino~\cite{tsounis2026kamino} at higher simulation throughput.
\end{itemize}
This extended abstract provides a brief account of preliminary research results that improve on Lodos-v1 \cite{beker2026lodos} that will eventually be incorporated into a full-length research paper.

\section{Background}\label{sec:problem}

\textbf{Notation:} At each simulation step, the unconstrained rigid-body dynamics routine computes the generalized inertia matrix $M \in \mathbb{R}^{n_v \times n_v}$ and free acceleration $\ddot q_f \in \mathbb{R}^{n_v}$, while the collision detection routine identifies $N$ contacts. Each contact $i$ is characterized by a local frame $(\hat n_i, \hat t_{1,i}, \hat t_{2,i})$, a Jacobian $J_i \in \mathbb{R}^{3\times n_v}$, a friction coefficient $\mu_i$, and a signed distance $d_i \le 0$.
We further denote the Delassus matrix as $G = JM^{-1}J^\trsp$, the free contact velocity as $v_f = J\dot q_f$, where $\dot q_f = \dot q + \ddot q_f\Delta t$, the product of elliptic friction cones as $\mathcal{K} = \prod_i \mathcal{K}_{\mu_i}$, with $\mathcal{K}_\mu = \{f : \|f_t\| \le \mu f_n\}$, the de~Saxc\'e correction~\cite{DeSaxceFeng1991} as $\Gamma_i(\sigma) = \mu_i\|\sigma_{i,t}\|\hat n_i$, and the penetration resolution term as $\gamma_i = \gamma_{n,i}\hat n_i$.

\textbf{The exact frictional contact problem:} The exact frictional contact problem refers to finding contact forces that satisfy the following nonlinear cone complementarity condition (N-CCP) \cite{carpentier2024compliant}:
\begin{align}
    \text{Find } f &: \nonumber \\
    \mathcal{K} &\ni f \perp (v^+ + \Gamma(v^+) + \gamma) \in \mathcal{K}^* \label{eq:nccp}
\end{align}
where $v^+ = v_f + Gf\Delta t$ is the end-of-step contact velocity and $\mathcal{K}^*$ is the dual friction cone. Given that $v^+$ is a linear function of $f$, an auxiliary velocity $\sigma = v^+$ can be introduced to decouple the nonlinear term $\Gamma(v^+)$ as $\Gamma(\sigma)$, resulting in an equivalent problem that only involves a linear cone complementarity condition:
\begin{align}
    \text{Find } f, \sigma &: \nonumber \\
    \sigma &= v_f + G f \Delta t, \label{eq:lccp_1} \\
    \mathcal{K} &\ni f \perp (v^+ + \Gamma(\sigma) + \gamma) \in \mathcal{K}^* \label{eq:lccp_2}
\end{align}
When $\sigma$ is fixed, \eqref{eq:lccp_2} is equivalent to the optimality condition of the following quadratically constrained quadratic program (QCQP):
\begin{equation}
    f^* = \argmin_{f\in\mathcal{K}}\ \tfrac{1}{2}f^\trsp (G  \Delta t) f + (v_f + \Gamma(\sigma) + \gamma)^\trsp f.
\end{equation}
Therefore, the exact nonlinear cone complementarity problem can be equivalently transcribed as~\cite{acary2018solving, carpentier2024compliant}:
\begin{align}
    \text{Find } f^*, \sigma &: \nonumber \\
    \sigma &= v_f + G f^* \Delta t, \label{eq:exact_1}  \\
    f^* &= \argmin_{f\in\mathcal{K}}\ \tfrac{1}{2}f^\trsp (G\Delta t) f + (v_f + \Gamma(\sigma) + \gamma)^\trsp f. \label{eq:exact_2}
\end{align}

\textbf{Rank Deficiency and Proximal Minimization:} Let us consider a generic convex optimization problem with a quadratic objective:
\begin{equation}
    \min_{x\in\mathcal{C}}\tfrac12 x^\trsp Q x + c^\trsp x
\end{equation}
The case when $Q$ is rank deficient is pathological, as there is no unique solution or $Q^{-1}$.
The \say{proximal minimization} method \cite{moreau1965proximite, rockafellar1976monotone} provides a general framework for addressing such cases:

\begin{algorithm}[H]
\caption{Proximal Minimization}
\label{alg:proximal-point}
\begin{algorithmic}[1]
\State Choose $\rho > 0$, $D \succ 0$, $x^0\in\mathcal C$; set $z \gets x^0$
\While{not converged}:
    \For{$k = 1, \dots, m$}
        \State Do an optimization update on
        \State $\min_{x\in\mathcal{C}}\tfrac12 x^\trsp Q x + c^\trsp x +\tfrac\rho2\|x-z\|_{D}^2$
        \State using any constrained optimization
        \State algorithm to obtain $x^*$.
    \EndFor
    \State $z \gets x^*$
\EndWhile
\State \Return $x^*$
\end{algorithmic}
\end{algorithm}
The algorithm repeatedly adds a quadratic penalty centered at the previous iterate $z$ and applies $m$ optimization updates to the resulting problem $\min_{x\in\mathcal{C}}\tfrac12 x^\trsp Q x + c^\trsp x +\tfrac\rho2\|x-z\|_{D}^2$. The penalty adds $\rho D \succ 0$ to the Hessian, so every inner problem is strongly convex. At a fixed point $z = x^*$, the penalty vanishes, so $x^*$ solves the original problem. The outer iteration converges to a minimizer even when $Q$ is only positive semi-definite, provided the inner problems are solved accurately enough \cite{rockafellar1976monotone}. The main benefit of the framework is its flexibility: any constrained optimization algorithm can be wrapped in an outer proximal minimization loop to improve the conditioning of its inner updates.

\textbf{Proximal Minimization for Frictional Contact:} Adding the proximal penalty $\tfrac\rho2\|f-z\|_{D}^2$ to the QCQP objective \eqref{eq:exact_2} and collecting linear and quadratic terms together gives
\begin{equation}\label{eq:socqp_prox}
    \min_{f \in \mathcal{K}} \: \frac{1}{2}f^\trsp ((G + \frac{\rho}{\Delta t}D) \Delta t)f + (v_f + \Gamma(\sigma) + \gamma - \rho Dz)^\trsp f.
\end{equation}
Let us define $R = \frac{\rho}{\Delta t}D$ and $r = v_f + \Gamma(\sigma) + \gamma - \rho Dz$ for notational brevity, and instead write:
\begin{equation}
    \min_{f\in\mathcal{K}}\tfrac12 f^\trsp((G + R)\Delta t)f + r^\trsp f \label{eq:dual}
\end{equation}
By cascading \cref{eq:exact_1} and \cref{eq:exact_2} into a fixed-point iteration scheme, and wrapping it in a proximal minimization loop, Carpentier et al.~\cite{carpentier2024compliant} propose the following solution routine for the exact frictional contact problem, as implemented in Kamino \cite{tsounis2026kamino}:
\begin{algorithm}[H]
\caption{Proximal ADMM with de Saxc\'e Correction}
\label{alg:proximal-point-correction}
\begin{algorithmic}[1]
\State Choose $\rho > 0$, diagonal $D \succ 0$, $R = \frac{\rho}{\Delta t}D$
\State Initialize $z\in\mathcal K$, $\sigma \gets v_f$
\While{not converged}:
    \State $r \gets v_f + \Gamma(\sigma) + \gamma - \rho Dz$
    \For{$k = 1, \dots, m$}
        \State Do an ADMM update on
        \State  $\min_{f\in \mathcal{K}} \: \frac{1}{2}f^\trsp  \left([(G + R)\Delta t]\right)f + r^\trsp f$
        \State to obtain $f^*$.
    \EndFor
    \State $z \gets f^*$
    \State $\sigma \gets v_f+Gf^*\Delta t$
\EndWhile
\State \Return $f^*$
\end{algorithmic}
\end{algorithm}
\section{Method: An Efficient Primal-Form Solver}
A transcription of the form \cref{eq:dual} is called the dual-form of contact dynamics, with contact forces as the decision variables. Whenever $R \succ 0$, the solution of \cref{eq:dual} is equivalent to the solution of the following nonlinear least-squares (NLS) transcription \cite{MuJoCo150Release}:
\begin{equation}
\min_{\ddot q}\ \tfrac12(\ddot q - \ddot q_f)^\trsp M(\ddot q - \ddot q_f)
+ \tfrac12\textstyle\sum_{i} R^{-1}_{ii}\|\Pi_{\mathcal{K}_i}(\arefi{i} - J_i\ddot q)\|^2,
\label{eq:nls}
\end{equation}
with $\aref = J\ddot q_f - r/\Delta t$, which is also called the primal-form of the contact dynamics, with generalized acceleration $\ddot q$ as the main decision variable. Lodos solves the exact frictional contact problem by altering \cref{alg:proximal-point-correction} so that it operates on the NLS transcription and uses Newton updates rather than ADMM updates:
\begin{algorithm}[H]
\caption{Proximal Newton with de Saxc\'e Correction}
\label{alg:proximal-point-correction-lodos}
\begin{algorithmic}[1]
\State Choose $\rho > 0$, diagonal $D \succ 0$, $R = \frac{\rho}{\Delta t}D$
\State Initialize $z\in\mathcal K$, $\sigma \gets v_f$
\While{not converged}:
    \State $r \gets v_f + \Gamma(\sigma) + \gamma - \rho Dz$
    \State $\aref \gets J\ddot q_f - r/\Delta t$
    \For{$k = 1, \dots, m$}
        \State Do a Newton update on
        \State $\min_{\ddot{q}} \: \frac{1}{2}(\ddot{q} - \ddot{q}_f)^\trsp M (\ddot{q} - \ddot{q}_f)$
        \State $\quad \quad \:\: + \frac{1}{2}\sum_{i=1}^N R^{-1}_{ii}\|\Pi_{\mathcal{K}_i}(a_\text{ref, i} - J_i \ddot{q})\|^2$
        \State to obtain $\ddot{q}^*$.
    \EndFor
    \State $f^* \gets R^{-1}\Pi_{\mathcal{K}}(\aref - J\ddot q^*)$
    \State $z \gets f^*$
    \State $\sigma \gets v_f+Gf^*\Delta t$
\EndWhile
\State \Return $f^*$
\end{algorithmic}
\end{algorithm}

Lodos-v2 supports two simulation modes. The first, the \say{exact mode}, implements \cref{alg:proximal-point-correction-lodos} to compute an exact solution to the frictional contact problem. The second, the \say{MuJoCo mode}, instead solves \cref{eq:nls}, but uses the definitions of $R$ and $a_\text{ref}$ from MuJoCo's formulation \cite{todorov2014convex}, rather than those arising from the proximal minimization penalty and the de Saxc\'e correction.

\section{Method: Contact-Group Reparameterization}
Lodos-v1~\cite{beker2026lodos} rescaled the spring-damper forces of a collision pair's contacts by one factor per
pair, $f = F\alpha$ with $F \in \mathbb{R}^{3N\times|\mathcal{P}|}$ over the collision pairs $\mathcal{P}$ and
${\alpha\succeq 0}$. Lodos-v2 generalizes this to \emph{contact groups}: a group $g \in \mathcal{G}$ is any subset of
all contacts of one collision pair, and one variable $\lambda_g \in
\mathbb{R}^{n_g}$ with $n_g \in \{1,3\}$ sets the forces of all group members. A group takes one of three parameterizations:
\begin{equationbox}{Contact-Group Reparameterization}
\begin{mysubequations}\label{eq:groups}
\begin{alignat}{2}
\text{per contact:}\enspace & f_i = C_i\lambda_i, &\enspace \lambda_i &\in \mathcal{K}_{\mu_i}, \label{eq:percontact}\\
\text{ray:}\enspace & f_i = a_i(\hat n_i - r_i\hat t_{1,i})\lambda_g, &\enspace \lambda_g &\geq 0, \label{eq:ray}\\
\text{cone:}\enspace & f_i = a_iC_i\lambda_g, &\enspace \lambda_g &\in \mathcal{K}_{\mu}. \label{eq:cone}
\end{alignat}
\end{mysubequations}
\end{equationbox}
For the per contact parameterization, no grouping is involved: each contact forms its own singleton group, such that $g=i$ and $n_g=3$. In contrast, the ray ($n_g=1$) and cone ($n_g=3$) parameterizations support arbitrary groupings of contacts. In practice, these groupings are constructed at each time step by binning contacts into a fixed number of groups according to the wrench contributions associated with their normal and tangential velocities. For each contact $i$ (independent of the grouping), the associated contact frame is $C_i = [\hat n_i\ \hat t_{1,i}\ \hat t_{2,i}] \in \mathbb{R}^{3\times 3}$, with $\hat t_{1,i}$ along its
tangential velocity $v_{i,t}$ and $\hat t_{2,i} = \hat n_i\times\hat t_{1,i}$. Its weight $a_i$ and (in the case of the ray parameterization) its friction ratio
$r_i$ come from the spring-damper model used by~\cite{beker2026lodos}, evaluated at the step's start and fixed for the step:
\begin{equation}
a_i = \frac{\bar f_{n,i}}{\sum_{j\in g}\bar f_{n,j}},\qquad \bar f_{n,i} \propto f_d\Big(\frac{v_{i,n}}{\delta}\Big)\max(-d_i, 0),
\label{eq:weights}
\end{equation}
with $f_d(x) = 1 - x$ for $x < 0$, $(x - 2)^2/4$ for $0 \le x < 2$ and $0$ beyond, and
$r_i = \mu\|v_{i,t}\|/(\epsilon^2 + \|v_{i,t}\|^2)^{1/2} < \mu$, a spring-damper model with dissipation and smooth stiction \cite{kurtz2026inverse}.

\textbf{The QCQP Corresponding to Each Parameterization:} Stacking the $\lambda_g$ of all groups into a single vector $\lambda \in \mathbb{R}^{N_g}$, $N_g = \sum_g n_g$, each parameterization forms a linear map
$f = B\lambda$ with $B \in \mathbb R^{3N \times N_g}$. Substituting this parameterization into~\eqref{eq:exact_2} preserves the exact same QCQP form, this time in terms of $\lambda$ rather than $f$:
\begin{align}
\sigma &= v_f + GB\lambda^*\Delta t, \nonumber\\
\lambda^* &= \argmin_{\lambda\in\tilde{\mathcal{K}}}\ \tfrac{1}{2}\lambda^\trsp\big((\tilde G + \tilde R)\Delta t\big)\lambda + \big(\tilde v_f + \tilde\Gamma(\sigma) + \tilde\gamma\big)^\trsp\lambda.
\label{eq:groupqcqp}
\end{align}
Here $\tilde G = B^\trsp GB = \tilde JM^{-1}\tilde J^\trsp \in \mathbb{R}^{N_g\times N_g}$, with the
groups' Jacobian $\tilde J = B^\trsp J \in \mathbb{R}^{N_g\times n_v}$, whose block for group $g$ is
$\tilde J_g = \sum_{i\in g} B_i^\trsp J_i \in \mathbb{R}^{n_g\times n_v}$;
$\tilde R = B^\trsp RB \in \mathbb{R}^{N_g\times N_g}$; $\tilde v_f = B^\trsp v_f$;
$\tilde\gamma = B^\trsp\gamma$, which puts $\sum_{i\in g} a_i\gamma_{n,i}$ on $\lambda_g$'s normal component; and
$\tilde{\mathcal{K}} = \prod_g\tilde{\mathcal{K}}_g$, with $\tilde{\mathcal{K}}_g$ the group's set
in~\eqref{eq:groups}. The problem overall has $3|\mathcal{G}| = 3N$ constraint rows for the per contact parameterization, $3|\mathcal{G}|$ for the cone parameterization, and
$|\mathcal{G}|$ for the ray parameterization (plus all other constraint rows from non-contact constraints). The de~Saxc\'e correction term, this time added to $\lambda_g$'s normal component, is constructed from contact velocities $\sigma_i$ as:
\begin{equation}
\tilde\Gamma_g(\sigma) = \begin{cases}
\mu_i\|\sigma_{i,t}\| & \text{per contact } (g = \{i\}),\\
\sum_{i\in g} a_ir_i\,\hat t_{1,i}^\trsp\sigma_i & \text{ray},\\
\mu\sum_{i\in g} a_i\|\sigma_{i,t}\| & \text{cone}.
\end{cases}
\label{eq:groupterm}
\end{equation}

\begin{algorithm*}[t]
\caption{MJWarp's Newton solver on the NLS transcription~\eqref{eq:nls}}
\label{alg:newton}
\begin{algorithmic}[1]
\Require rows $(J, R, \aref)$, cone $\mathcal{K}$, $M$, $\ddot q_f$; warm start $\ddot q_0$;
 $\varepsilon$ = \code{tolerance}, $k_{\max}$ = \code{iterations}, $\ell_{\max}$ = \code{ls_iterations},
 $\varepsilon_{\ell}$ = \code{ls_tolerance}
\State $\ddot q \gets \ddot q_0$ \Comment{last step's $\ddot q$ with \code{warm_start}, else $\ddot q_f$}
\State $y \gets \aref - J\ddot q$;\ \ $f \gets R^{-1}\Pi_{\mathcal{K}}(y)$;\ \ each row's state: inactive, quadratic, or the cone's middle zone
\State $g \gets M(\ddot q - \ddot q_f) - J^\trsp f$
\State $H \gets M + J^\trsp R^{-1}\,\partial\Pi_{\mathcal{K}}(y)\,J \in \mathbb{R}^{n_v\times n_v}$ \Comment{\code{shared_hessian}: summed per world in shared memory}
\State $d \gets -H^{-1}g$ \Comment{Cholesky; \code{row_space}: through the rows, $H$ never formed}
\For{$k = 1, \dots, k_{\max}$} \Comment{while any world is unconverged}
  \State $\alpha \gets$ exact line search on $\ell(\ddot q + \alpha d)$, the objective~\eqref{eq:nls}: bracket refinement
  \Statex \hskip\algorithmicindent until $\lvert \partial\ell/\partial\alpha\rvert < \max(\varepsilon\,\varepsilon_\ell\,\lVert d\rVert\, s,\ 10^{-6})$, at most $\ell_{\max}$ steps
  \State $\ddot q \gets \ddot q + \alpha d$; recompute $y$, $f$, the row states, $g$, $H$ and $d$ (lines 2--5)
  \If{$(\ell_{\mathrm{prev}} - \ell)/s < \varepsilon$ \textbf{or} $\lVert g\rVert/s < \varepsilon$ \textbf{or} $\tfrac12 g^\trsp H^{-1}g/s < \varepsilon$}
    \State freeze the world: converged
  \EndIf
\EndFor \Comment{a world still running at $k_{\max}$ freezes with the \code{ITERATIONS} flag}
\State \Return $\ddot q$, $f$
\end{algorithmic}
\end{algorithm*}

\section{Method: Implementation}\label{sec:newton}
The MJWarp codebase~\cite{mujoco_warp} provides an excellent, highly optimized unconstrained Newton solver for problems of the form given in \cref{eq:nls}. This solver, the nonsmooth Newton method of Todorov~\cite{MuJoCo150Release}, is described in \Cref{alg:newton} exactly as MJWarp implements it, with the same line search and convergence test. In MJWarp (and by extension in Lodos-v2), every simulation instance in a batched simulation runs \Cref{alg:newton} independently in the same kernels. A converged world freezes and skips all later work, and the loop is a device-side while (a CUDA graph conditional node) over the unconverged worlds.
Lodos-v2 builds upon MJWarp's solver as its computational core. Three new options change how lines 4--5 are computed, as discussed next.
\\
\\
\textbf{Row Space}: Lodos-v1 solved the dual form problem \eqref{eq:dual}, and reduced its number of decision variables from $3N$ (i.e. three times the number of active contacts) to $K$ (i.e. the number of active collision pairs) using contact-group reparameterization. This is still more efficient than solving the primal form whenever $n_v \geq K$ (e.g. piles of free-bodies like a wall of bricks), but it does not readily lend itself into a combination with other types of constraints. The row-space option addresses this problem by formulating the problem in primal form, but using Woodbury identity to factor the corresponding Hessian in row-space (a matrix of size $K \times K$) rather than in generalized acceleration space (a matrix of size $n_v \times n_v$).
Let $M = LL^\trsp,\ \ L \in \mathbb{R}^{n_v\times n_v}$ be the Cholesky factor of the mass matrix, lower triangular and block diagonal over the kinematic trees, and $W^\trsp W = R^{-1}\partial\Pi_{\mathcal{K}}(y),\ \ W \in \mathbb{R}^{m\times 3N}$ be the Cholesky factor of the constraint-space Hessian (in line~4 of \cref{alg:newton}), whose $m$ rows now include the $K$ constraints from Lodos-v1's contact-group reparameterization as well as other constraint types from MuJoCo. The Hessian $H$ factors through $L$, and the Woodbury identity inverts its middle factor, which gives $H^{-1}g$ as:
\begin{align}
M &= LL^\trsp,\ \ L \in \mathbb{R}^{n_v\times n_v}, \nonumber \\
G &= Z^\trsp Z,\ Z = L^{-1}J^\trsp \in \mathbb{R}^{n_v\times 3N}, \nonumber\\
W^\trsp W &= R^{-1}\partial\Pi_{\mathcal{K}}(y),\ \ W \in \mathbb{R}^{m\times 3N}, \nonumber\\
H &= M + J^\trsp W^\trsp WJ = L\big(\Id_{n_v} + ZW^\trsp WZ^\trsp\big)L^\trsp, \nonumber\\
\nonumber \\
H^{-1}g &= L^{-\trsp}\big(\eta - ZW^\trsp\xi\big), \nonumber\\
\eta &= L^{-1}g,\quad \xi = (\Id_m + WGW^\trsp)^{-1}WZ^\trsp\eta,
\label{eq:rowspace}
\end{align}
\\
\textbf{Shared Hessian}: Sums the sparse $H$ per world in the shared memory of each streaming multiprocessor (SM) \cite{nvidia_cuda_guide}, if it fits.
\\
\textbf{Fused Newton}: Runs \cref{alg:newton} and the proximal rounds around it as one kernel per world, therefore storing all the data (e.g. the solver state) in the shared memory of the SM, if it fits.
\\
With these improvements, \Cref{alg:exact} and \Cref{alg:mj} describe the solution routines implemented by each mode of Lodos.

\section{Results}\label{sec:remarks}
\begin{table*}[t]
\centering\small\setlength{\tabcolsep}{6pt}
\begin{tabular}{@{}lrrrrr@{}}
\toprule
 & \multicolumn{3}{c}{Lodos-v2, mj mode} & & \\
\cmidrule(lr){2-4}
scene & per contact & cone & ray & MJWarp & speed-up \\
\midrule
\texttt{cube\_walled\_5\_polytope} & 1.61M @8192 & -- & -- & 1.11M @8192 & 1.45$\times$ \\
\texttt{cube\_walled\_5\_sdf} & 551k @4096 & 1.27M @8192 & 1.86M @8192 & 276k @2048 & 6.76$\times$ \\
\texttt{arm\_walled\_5\_polytope} & 985k @8192 & -- & -- & 163k @2048 & 6.04$\times$ \\
\texttt{arm\_walled\_5\_sdf} & 1.24M @4096 & 1.25M @8192 & 1.97M @8192 & 238k @2048 & 8.29$\times$ \\
\texttt{shadow\_grasp\_polytope} & 1.18M @8192 & -- & -- & 432k @32768 & 2.73$\times$ \\
\texttt{shadow\_grasp\_sdf} & 1.38M @8192 & 1.25M @8192 & 1.42M @8192 & 126k @4096 & 11.27$\times$ \\
\texttt{shadow\_walled\_5\_polytope} & 130k @8192 & -- & -- & 36.2k @2048 & 3.60$\times$ \\
\texttt{shadow\_walled\_5\_sdf} & 112k @4096 & 113k @8192 & 110k @8192 & 17.4k @4096 & 6.46$\times$ \\
\texttt{nist\_threading\_m8} & 1.74M @8192 & 1.94M @8192 & 1.97M @8192 & 52k @1024 & 37.35$\times$ \\
\texttt{cube\_walled\_20\_polytope} & 394k @16384 & -- & -- & 256k @16384 & 1.54$\times$ \\
\texttt{cube\_walled\_20\_sdf} & 119k @4096 & 263k @16384 & 349k @8192 & 107k @4096 & 3.25$\times$ \\
\texttt{arm\_walled\_20\_polytope} & 178k @16384 & -- & -- & 46.3k @2048 & 3.85$\times$ \\
\texttt{arm\_walled\_20\_sdf} & 261k @16384 & 259k @16384 & 366k @16384 & 84k @4096 & 4.36$\times$ \\
\texttt{shadow\_walled\_20\_polytope} & 3.31k @512 & -- & -- & 2.59k @512 & 1.28$\times$ \\
\texttt{shadow\_walled\_20\_sdf} & 2.88k @512 & 2.3k @512 & 2.36k @512 & 626 @64 & 4.60$\times$ \\
\bottomrule
\end{tabular}
\vspace{5pt}
\caption{\captitle{Lodos-v2's MuJoCo mode against MJWarp.} Each configuration's best throughput, in world-steps per
second at the batch size after @ (k $= 10^3$, M $= 10^6$), and the speed-up of Lodos-v2's fastest configuration over MJWarp's best.}
\label{tab:mjwarp}
\end{table*}

\begin{table*}[t]
\centering\small\setlength{\tabcolsep}{6pt}
\begin{tabular}{@{}lrrrrr@{}}
\toprule
 & \multicolumn{3}{c}{Lodos-v2, exact mode} & & \\
\cmidrule(lr){2-4}
scene & per contact & cone & ray & Kamino & speed-up \\
\midrule
\texttt{cube\_walled\_5\_polytope} & 711k @8192 & -- & -- & 589k @4096 & 1.21$\times$ \\
\texttt{cube\_walled\_5\_sdf} & 225k @2048 & 596k @8192 & 1.08M @4096 & -- & 1.83$\times^{\ddagger}$ \\
\texttt{arm\_walled\_5\_polytope} & 408k @8192 & -- & -- & 58.3k @128 & 6.99$\times$ \\
\texttt{arm\_walled\_5\_sdf} & 523k @4096 & 544k @4096 & 1.12M @8192 & -- & 19.26$\times^{\ddagger}$ \\
\texttt{shadow\_grasp\_polytope} & 983k @8192 & -- & -- & 11.1k @512 & 88.33$\times$ \\
\texttt{shadow\_grasp\_sdf} & 660k @8192 & 675k @8192 & 1.02M @8192 & -- & 60.64$\times^{\ddagger}$ \\
\texttt{shadow\_walled\_5\_polytope} & 103k @8192 & -- & -- & 501 @256 & 205.50$\times$ \\
\texttt{shadow\_walled\_5\_sdf} & 65.3k @4096 & 69.4k @8192 & 74.4k @4096 & -- & 148.50$\times^{\ddagger}$ \\
\texttt{nist\_threading\_m8} & 1.12M @8192 & 1.25M @8192 & 1.45M @8192 & -- & -- \\
\texttt{masonry\_arch\_polytope} & 175k @16384 & -- & -- & 35.3k @2048 & 4.97$\times$ \\
\texttt{masonry\_arch\_sdf} & 44.3k @2048 & 34.8k @1024 & 21.4k @1024 & -- & 1.26$\times^{\ddagger}$ \\
\texttt{house\_of\_cards\_polytope} & 34.2k @8192 & -- & -- & 9.34k @512 & 3.66$\times$ \\
\texttt{house\_of\_cards\_sdf} & 7.89k @2048 & 18.6k @4096 & 16.8k @2048 & -- & 1.99$\times^{\ddagger}$ \\
\texttt{cube\_walled\_20\_polytope} & 189k @16384 & -- & -- & 76.6k @4096 & 2.47$\times$ \\
\texttt{cube\_walled\_20\_sdf} & 61k @4096 & 148k @16384 & 199k @8192 & -- & 2.60$\times^{\ddagger}$ \\
\texttt{arm\_walled\_20\_polytope} & 102k @16384 & -- & -- & 9.66k @256 & 10.59$\times$ \\
\texttt{arm\_walled\_20\_sdf} & 143k @16384 & 142k @16384 & 216k @16384 & -- & 22.31$\times^{\ddagger}$ \\
\texttt{shadow\_walled\_20\_polytope} & 2.62k @1024 & -- & -- & left out & -- \\
\texttt{shadow\_walled\_20\_sdf} & 1.74k @512 & 1.58k @512 & 1.56k @512 & -- & -- \\
\bottomrule
\end{tabular}
\vspace{5pt}
\caption{\captitle{Lodos-v2's exact mode against Kamino.} The best throughput achieved by each configuration and the speed-up of Lodos-v2's fastest configuration. Kamino is evaluated only on the polytope versions of the scenes. The symbol $^\ddagger$ indicates comparisons between Lodos-v2 on the original surfaces and Kamino on the corresponding polytope versions. The threading task has no polytope version, and therefore does not include a Kamino baseline. Similarly, Kamino was excluded from the twenty-hands scene because benchmarking was prohibitively slow (at a batch size of 16, each simulation step took 7.2\,s).}
\label{tab:kamino}
\end{table*}

This section summarizes benchmark experiments comparing Lodos-v2 with those solvers from the Newton~\cite{The_Newton_Contributors_Newton_GPU-accelerated_physics_2025} framework\footnote{particularly the 1.7.0 version.} whose contact dynamics its two modes reproduce: the MuJoCo mode against MJWarp (Newton's \code{SolverMuJoCo}), and the exact mode
against Kamino~\cite{tsounis2026kamino} (Newton's \code{SolverKamino}). The main comparison metric is batched throughput, in
world-steps per second, on one NVIDIA RTX~4080 with 16\,GB of memory. \Cref{tab:mjwarp,tab:kamino} give each
configuration's best throughput and the speed-up of Lodos-v2 over each baseline, which \cref{fig:teaser} plots
under the scenes. The exact experimental procedure, scenes, and settings can be found in the associated code release, which also provides utilities to reproduce the benchmark.
\\
\\
\textbf{Scenes.} Comparisons use the following scenes:
\begin{itemize}[leftmargin=*, label={\color{bulletcolor}\textbf{\textbullet}}]
    \item \code{cube_walled_5/20}, \code{arm_walled_5/20}, \code{shadow_walled_5/20}: Scenes containing piles of 5/20 cubes, armadillos, or Shadow hands, respectively, dropped into a walled box. The cube scenes evaluate performance on free-floating rigid bodies with simple convex geometry, the armadillo scenes evaluate performance on free-floating rigid bodies with detailed nonconvex geometry, and the Shadow hand scenes evaluate performance on complex kinematic trees with detailed nonconvex geometry. Kamino is specifically designed for accurate simulation of kinematic structures with closed-loop topologies \cite{tsounis2026kamino}, a specialization that inherently limits its performance on kinematic trees. Accordingly, the cube and armadillo scenes provide a comparison between Kamino and Lodos-v2 under equivalent kinematic constraints, as neither involves kinematic trees or closed-loop structures. The Shadow hand scenes, by contrast, feature complex kinematic trees and are intended to highlight the complementary specializations of the two frameworks: Lodos-v2's focus on kinematic trees and Kamino's focus on closed-loop structures.
    \item \code{shadow_grasp}: a Shadow hand lifting a toy, replaying the position command from a prerecorded grasp. The scene replays an open-loop control sequence that is squeezing a delicate, four-fingered prismatic grasp. This means maintaining a stable grasp without slipping requires highly accurate friction modeling.
    \item \code{nist_threading_m8}: A Franka arm threads an M8 nut onto a bolt by replaying an open-loop control sequence recorded from an RL policy \cite{zhang2026balanced}. The fine screw threads of the nut and bolt require an accurate representation of detailed, non-convex collision geometry, making this a challenging simulation scenario.
    \item \code{masonry_arch}: A masonry arch composed of 25 stones~\cite{SongFAP2026Splitting}. Maintaining the arch's stability requires physically accurate modeling of frictional interactions between the stones.
    \item \code{house_of_cards}: A house of 40 cards~\cite{SongFAP2026Splitting}. It likewise requires accurately modeling frictional interactions between the cards.
\end{itemize}
Each scene but \code{nist_threading_m8} comes in two versions, sharing the same bodies, inertias and
starts. For the \code{_sdf} version, Lodos-v2 collides the bodies using the collision pipeline of Lodos-v1 (i.e. vertices against signed distance fields), and Newton using its mesh-SDFs collision pipeline (with contact reduction on). For the \code{_polytope} version, both collide them as polytopes obtained from convex decomposition \cite{wei2022coacd}.
The threading scene only has the \code{_sdf} version as a convex decomposition does not provide adequate geometric fidelity given the fine screw threads. The Kamino baseline is evaluated only on the \code{_polytope} variants, as it was prohibitively slow to benchmark with the large number of contact points generated by Newton's mesh-SDF collision pipeline.
\\
\\
\textbf{Settings.} Lodos-v2 runs each \code{_sdf} scene under all three contact-group parameterizations (i.e. per contact, cone and ray), and each \code{_polytope} scene only using the per contact parameterization.  In the MJWarp comparison, both simulators use the same contact model parameters (solref 5\,ms with damping ratio 1; 2.5\,ms on the threading), the same parameters for the Newton routine (100 iterations,
tolerance $10^{-6}$, 50 line-search iterations), elliptic cones and the implicitfast
integrator. In the Kamino comparison, Lodos-v2's exact mode performs $n = 2$ proximal rounds with $\hat\rho = 0.05$ and $\alpha_B = 0.05$ (\cref{alg:exact}), while Kamino uses its default Newton solver settings with the number of iterations capped at 20.\footnote{For the hand scenes, the cap is increased to 200 iterations, as fewer iterations resulted in physical artifacts.} These settings are chosen such that both solvers pass the physical faithfulness benchmark of \cite{SongFAP2026Splitting}. This benchmark comprises scenes specifically designed to expose the limitations of simulators based on convex relaxations (such as MuJoCo), which produce physical artifacts, while requiring simulators that solve the exact frictional contact problem to achieve physically faithful results.
\\
\\
\textbf{Measurement.} Every world in a batch starts from its own unique state to eliminate confounders due to caching: a pile from a random stack of its own, the other scenes
from their start with each free body randomly perturbed by up to 0.05\,mm. Each configuration's memory capacities (e.g. for the pre-allocated arrays that store contact and collision data) are 1.5 times a world's largest demand (over 16 randomized starts) times the batch size. A run times the whole simulated span after its step has been captured: 3\,s of a pile (the minimum time needed for the pile to settle), 5\,s of the grasp
and the threading (the duration of the pre-recorded control trajectories), 2\,s of the arch and the house of cards (the minimum time needed for the scene to settle). The batch size doubles from 64 worlds until the throughput drops or the batch no longer fits.
\\
\\
\textbf{Reproducing and viewing the results.} The code for running both benchmarks can be found in the folder \code{bench/} of the associated codebase (under \code{bench/lodos_mjwarp} and
\code{bench/lodos_kamino}). The file \code{bench/README.md} describes the steps needed to reproduce every number in
\cref{tab:mjwarp,tab:kamino}, and each benchmark's report gives the full
settings, the capacities, the execution options and the throughput at every batch size. The file \code{viewer/README.md}
shows how to open each scene in the live viewer that comes with the codebase, under all three simulators.

\section{Conclusion}
In this extended abstract, we have presented Lodos-v2, a framework for high-throughput simulation of rigid-body dynamics with contact in scenes involving large numbers of collision pairs and detailed geometries. An immediate direction for future work is to demonstrate the framework's potential to accelerate applications such as reinforcement learning (RL), model predictive control (MPC), and randomized smoothing.

\newpage

\begin{algorithm*}[t]
\caption{The exact mode of Lodos: proximal minimization with de~Saxc\'e correction around \cref{alg:newton}}
\label{alg:exact}
\begin{algorithmic}[1]
\Require $\hat\rho$ = \code{prox_rho}, $n$ = \code{prox_iters_per_qp} $\ge 1$, $\varepsilon_z$ = \code{prox_tol_z},
 $\varepsilon_\sigma$ = \code{prox_tol_sigma}, $\alpha_B$ = \code{baumgarte_alpha}, $s_B$ = \code{baumgarte_slop}
\State $\mathcal{K} \gets$ the elliptic cones; impratio $\gets 1$
\For{each contact $i$} \Comment{\code{_prox_init}, all contacts in parallel}
  \State $R_{ii} \gets \hat\rho\, w_i$
  \State $\gamma_{n,i} \gets \alpha_B \min\bigl(0, (d_i + s_B)/\Delta t\bigr) + \max(0, d_i/\Delta t)$ \Comment{$0$ without \code{baumgarte}}
  \State $z_i \gets \Pi_{\mathcal{K}_i}(\text{its cached anchor})$ if \code{warm_start} matches it in the cross-step cache, else $0$
  \State $\sigma_i \gets J_i(\dot q + \Delta t\,\ddot q_0)$ \Comment{at the warm start}
\EndFor
\State $(n_o, n_i) \gets (n, 1)$ with \code{refresh_de_saxce_per_prox}, else $(1, n)$
\For{$o = 1, \dots, n_o$}\label{line:rounds} \Comment{each round after the first is skipped once every world is frozen}
  \State $\Gamma_i \gets \mu_i\lVert\sigma_{i,t}\rVert\,\hat n_i$ \Comment{$0$ without \code{de_saxce}}
  \For{$j = 1, \dots, n_i$}
    \State $\arefi{i,n} \gets -\bigl(v_{i,n} + \gamma_{n,i} + \Gamma_{i,n}\bigr)/\Delta t + R_{ii} z_{i,n}$;\ \
           $\arefi{i,t} \gets -v_{i,t}/\Delta t + R_{ii} z_{i,t}$ \Comment{$v_i = J_i\dot q$}
    \State $\ddot q \gets$ \cref{alg:newton} on these rows, from the current $\ddot q$
    \State $f^* \gets R^{-1}\Pi_{\mathcal{K}}(\aref - J\ddot q)$;\ \ $\Delta z \gets \max\lvert f^* - z\rvert$;\ \ $z \gets f^*$
    \If{$n_i > 1$ \textbf{and} $\Delta z \le \varepsilon_z(1 + \max\lvert f^*\rvert)$} freeze the world \EndIf
  \EndFor
  \If{\code{refresh_de_saxce_per_prox}}
    \State $\sigma \gets J(\dot q + \Delta t\,\ddot q)$;\ \ $\Delta\Gamma \gets \max_i\lvert\Gamma_i(\sigma) - \Gamma_i\rvert$
    \If{$\Delta z \le \varepsilon_z(1 + \max\lvert f^*\rvert)$ \textbf{and} $\Delta\Gamma \le \varepsilon_\sigma(1 + \max\lvert\Gamma\rvert)$} freeze the world \EndIf
  \EndIf
\EndFor\label{line:endrounds}
\State with \code{warm_start}: write each contact's anchor into the cross-step cache (\code{_prox_cache_fill})
\State \Return $\ddot q$, $f^*$
\end{algorithmic}
\end{algorithm*}

\begin{algorithm*}[t]
\caption{The MuJoCo mode of Lodos}
\label{alg:mj}
\begin{algorithmic}[1]
\Require the model's cone and impratio; each row's \code{solref}, \code{solimp}; $\hat\rho$, $n$ = \code{prox_iters_per_qp}, \code{de_saxce}
\For{each contact $i$}
  \State $R_{ii} \gets w_i\,(1 - \mathrm{imp}_i)/\mathrm{imp}_i$, its friction rows' divided by impratio
  \State $\arefi{i} \gets k_i\,\mathrm{imp}_i\,\lvert d_i\rvert\,\hat n_i - b_i\,v_i$ \Comment{MuJoCo's spring-damper reference}
\EndFor
\If{$n = 0$ \textbf{and not} \code{de_saxce}} \Comment{MuJoCo's problem; nothing overridden}
  \State $\ddot q \gets$ \cref{alg:newton} on these rows, on the model's cone \Comment{elliptic: $\mu_r = \mu/\sqrt{\text{impratio}}$}
\Else \Comment{the proximal path on MuJoCo's soft rows (\code{opt.prox_soft}), elliptic cone}
  \State $\rho_i \gets \hat\rho\,w_i$ if $n > 0$, else $0$ \Comment{without a loop: one round, carrying the term alone}
  \State in \cref{alg:exact}, lines~\ref{line:rounds}--\ref{line:endrounds} with $R'_{ii} \gets R_{ii} + \rho_i$ and the references
  \Statex \hskip\algorithmicindent $\arefp{i,n} \gets \arefi{i,n} - b_i\,\Gamma_{i,n} + \rho_i z_{i,n}$,\ \ $\arefp{i,t} \gets \arefi{i,t} + \rho_i z_{i,t}$, and no Baumgarte bias
\EndIf
\State \Return $\ddot q$, $f$
\end{algorithmic}
\end{algorithm*}

\bibliographystyle{unsrtnat_links}
{\hypersetup{urlcolor=biblinkcolor}\bibliography{references}}

\makeatletter\let\baselinestretch\ACM@origbaselinestretch\makeatother
\end{document}